\documentclass[review]{elsarticle}

\usepackage{lineno,hyperref}
\usepackage{subfig}
\usepackage{amsmath}
\usepackage{algorithm}
\usepackage[noend]{algpseudocode}
\usepackage{adjustbox}
\usepackage{array}
\usepackage{multirow}
\usepackage{booktabs}

\journal{Journal of Artificial intelligence}

\begin{document}

\begin{frontmatter}

\title{Explaining Anomalies Detected by Autoencoders Using SHAP}

\author{Liat Antwarg\fnref{myfootnote}}
\author{Ronnie Mindlin Miller\fnref{myfootnote}}
\author{Bracha Shapira, Lior Rokach}
\address{Ben-Gurion University of the Negev, Israel}
\address{liatant@post.bgu.ac.il, Ronniemi@post.bgu.ac.il, bshapira@bgu.ac.il, liorrk@post.bgu.ac.il}
\fntext[myfootnote]{Equal contribution.}
\begin{abstract}
Deep learning algorithms for anomaly detection, such as autoencoders, point out the outliers, saving experts the time-consuming task of examining normal cases in order to find anomalies. Most outlier detection algorithms output a score for each instance in the database. The top-k most intense outliers are returned to the user for further inspection; however, the manual validation of results becomes challenging without justification or additional clues. An explanation of why an instance is anomalous enables the experts to focus their investigation on the most important anomalies and may increase their trust in the algorithm.
Recently, a game theory-based framework known as SHapley Additive exPlanations (SHAP) was shown to be effective in explaining various supervised learning models. 

In this paper, we propose a method that uses kernel SHAP to explain anomalies detected by an autoencoder, which is an unsupervised model. The proposed explanation method aims to provide a comprehensive explanation to the experts by focusing on the connection between the features with high reconstruction error and the features that are most important in affecting the reconstruction error. We propose a black-box explanation method, because it has the advantage of being able to explain any autoencoder without being aware of the exact model. The proposed explanation method extracts and visually depicts both the features that most contribute to the anomaly and those that offset it. A user study using real-world data demonstrates the usefulness of the proposed method in helping domain experts better understand the anomalies. Evaluation of the explanation method using a "perfect" autoencoder as the ground truth shows that the proposed method explains anomalies correctly, using the exact features, and evaluation on real-data demonstrates that (1) our explanation model which uses SHAP is more robust than LIME, and (2) the explanations our method provides are more effective at reducing the anomaly score than other methods.
\end{abstract}

\begin{keyword}
Explainable black-box models\sep autoencoder\sep shapley values \sep anomaly detection\sep XAI
\MSC[2010] 00-01\sep  99-00
\end{keyword}

\end{frontmatter}

\section{Introduction}
Recently, deep learning algorithms have been used for a wide variety of problems, including anomaly detection. While anomaly detection algorithms may be effective at identifying anomalies that might otherwise not be found, they have a major drawback, as their output is hard to explain. This shortcoming can make it challenging to convince experts to trust and adopt potentially beneficial anomaly detection systems. The output of such algorithms may contain anomalous instances that the domain expert was previously unaware of, and an explanation of why an instance is anomalous can increase the domain expert's trust in the algorithm. In addition, explanations can be contrastive, which is helpful and extremely relevant to explaining anomalies, since people do not usually ask why an event happened, but rather why one event happened instead of another event \cite{miller2018explanation}. 
The need to provide an explanation per instance (as opposed to providing an explanation for the whole model) was discussed in the 1970s in relation to expert systems \cite{shortliffe1975model}, but it has come to the fore again more recently, as models have become more complex. 
In some domains, such as the autonomous car and medical domains, the lack of understanding and validating the decision process of a machine learning system is a disadvantage. The absence of a proper explanation, makes it more difficult for decision-makers and domain experts to use the output of such algorithms.
Reliable explanations build trust with users, help identify points of model failure, and remove barriers to entry for the deployment of deep neural networks in different domains. By building more transparent, explainable systems, users will better understand and therefore trust the intelligent agents \cite{ribeiroshould, miller2018explanation, kindermans2017reliability}. 

In the last decade, a few methods have been developed to explain predictions from supervised models. One approach uses an interpretable approximation of the original model \cite{lundberg2017unified}; examples of methods that use an interpretable approximation of the original model include: LIME \cite{ribeiro2016should}, which is an example of a model-agnostic method used to explain a prediction using a local model, and DeepLIFT \cite{shrikumar2017learning}, which is an example of a model-specific method for explaining deep learning models in which the contributions of all neurons in the network are back-propagated to the input features. SHAP - SHapley Additive exPlanation \cite{lundberg2017unified} combines previous methods for explaining predictions by calculating feature importance, using Shapley values from game theory that ensures consistency of the explanations.

Recently, autoencoders have become widely adopted for unsupervised anomaly detection tasks \cite{erfani2016high,paula2016deep,sakurada2014anomaly}. An autoencoder is an unsupervised algorithm that represents the normal data in lower dimensionality and then reconstructs the data in the original dimensionality; thus, the normal instances are reconstructed properly, and the outliers are not, making the anomalies clear. To the best of our knowledge, no previous research has been performed to provide black-box explanation to anomalies revealed by an autoencoder. 
In this paper, we present a new method based on SHAP values, to explain the anomalies found in an autoencoder's output. The method could be beneficial to experts requiring justification and visualization regarding anomalies. Domain experts involved in our preliminary experiment on real-world data provided positive feedback, claiming that the explanations helped them understand and investigate the anomalies. In addition, to evaluate the proposed method, we created autoencoders for which we know the connections between the features; we then used an artificial dataset to examine whether our method uses the correct set of features to explain the anomalies. We also evaluated the robustness of our method using 'noise' features and investigated how the explanations can be used to reduce the anomaly score of an instance.

The contributions of our work to the field are as follows: (1) We developed a method to explain anomalies revealed by an unsupervised autoencoder model. The method explains the features with the highest reconstruction errors using Shapley values. This method could have significant practical value, since autoencoders have become a popular choice for anomaly detection in recent years. To the best of our knowledge, no other research has used a model-agnostic method to explain anomalies revealed by an autoencoder. (2) We conducted a preliminary experiment with real-world data and domain experts as part of a project aimed at improving the cost monitoring process of a large car manufacturer, to obtain feedback on the anomaly explanations provided by our method. (3) We suggested methods for evaluating the explanations. Since the approach of explaining instances (as opposed to the whole model) is quite new, and the explanation goals can be different in the two approaches, it is still not clear how to evaluate methods for explaining instances. In this research, we adopted three ideas used in previous studies on explanation methods and adjusted them to evaluate our model-agnostic method of explaining anomalous instances. We evaluated the correctness, robustness, and effectiveness of the explanations in reducing the anomaly score.  \newline

\section{Background}
\subsection{Autoencoders}
Autoencoders were first introduced in the 1980s \cite{rumelhart1985learning}, and in the last decade they have been widely used in deep architectures \cite{bengio2007scaling,hinton2006reducing,hinton2006fast}. An autoencoder is an unsupervised neural network that is trained to produce target values equal to its inputs \cite{goodfellow2016deep}. 
An autoencoder represents the data in lower dimensionality (encoding) and reconstructs the data into the original dimensionality (decoding). Based on the input, the autoencoder learns an identity function so that the autoencoder's output is similar to the input and the embedded model created in encoding represents normal instances well. In contrast, anomalies are not reconstructed well and have a high amount of reconstruction error, so in the process of encoding and decoding the instances, the anomalies are discovered.\newline
 \subsection{SHAP}
A few methods have been developed to explain predictions from supervised models. The SHAP framework \cite{lundberg2017unified} (SHapley Additive exPlanation) gathers previously proposed explanation methods, such as LIME \cite{ribeiro2016should} and DeepLIFT \cite{shrikumar2017learning}, under the class of additive feature attribution methods. Methods in that class are explanation models in the form of a linear function of simplified binary variables, as in $f(x)=g(z) = \theta_0 + \sum_{i=1}^{m}\theta_i z_i$, where $f(x)$ is the original model (an autoencoder in this paper); $g(x)$ is the explanation model; $z$ is the simplified input; $x=h_x(z)$ is a mapping function to the original method; and $\theta_0=f(h_x(0)$ is the model output without all of the simplified inputs. 

SHAP has a sound theoretic basis, which is a benefit in regulated scenarios. It uses Shapley values from game theory to explain a specific prediction by assigning an importance value (SHAP value) to each feature that has the following properties: (1) local accuracy - the explanation model has to at least match the output of original model; (2) missingness - features missing in the original input must have no impact; (3) consistency - if we revise a model such that it depends more on a certain feature, then the importance of that feature should not decrease, regardless of other features.

Lundberg and Lee \cite{lundberg2017unified} demonstrate that SHAP is better aligned with human intuition than previous methods, since it has those properties. The SHAP framework suggests a model-agnostic approximation for SHAP values, called Kernel SHAP. Kernel SHAP uses linear LIME \cite{ribeiro2016should}, along with Shapley values, to build a local explanation model. A local model uses a small background set from the data to build an interpretable model that takes the proximity to the instance to be explained into account \cite{ribeiro2016should}. We use Kernel SHAP as it provides more accurate estimates with fewer evaluations of the original model than other sampling-based estimates.
\section{Related work}
\subsection{Explaining unsupervised models}
Clustering is widely used for unsupervised learning problems. There has been limited success in addressing the issue of cluster interpretability \cite{bertsimas2018interpretable}. A popular explanation method is the representation of a cluster of points by their centroid or by a set of distant points in the cluster\cite{radev2004centroid}. This works well when the clusters are compact or isotropic but fails otherwise. 
Another common approach is the visualization of clusters in a two-dimensional graph using principle component analysis (PCA) projections or t-SNE \cite{jolliffe2011principal, maaten2008visualizing}. However, reducing the dimensionality of the features obscures the relationship between the clusters and the original variables.
Another way of explaining clustering is by building a decision tree for each cluster after the clustering process has taken place. In \cite{bertsimas2017optimal}, the authors took the opposite approach, proposing Interpretable Clustering via Optimal Trees (ICOT), in which they build a decision tree using the feature values (unsupervised), and the leaves are the clusters. Other studies also presented clustering methods using decision trees \cite{liu2000clustering, lundberg2018consistent}, in both cases building explainable clusters, rather than explaining clusters created using traditional clustering algorithms. Kauffman et al. \cite{kauffmann2018towards} suggest a deep Taylor decomposition of one-class SVM that provides an explanation using support vectors or input features. 

\subsection{Explaining anomalies}
According to Amarasinghe et al. \cite{amarasinghe2018toward}, the ability to explain an anomaly detection model in critical areas, such as infrastructure security, is almost as important as the model's prediction accuracy. Thus, effective anomaly explanation would significantly improve the usability
of anomaly detection techniques in real-world applications. One of the motivations for explaining anomalies revealed by an anomaly detection algorithm is to bridge the gap between detecting outliers and identifying domain-specific anomalies. In other words, some of the outliers are actually interesting anomalies and others are revealed by the algorithm as outliers only because they are rare (and not interesting as anomalies). Providing reasons for outlierness can significantly reduce the manual inspection performed by domain experts \cite{liu2018contextual}.

Liu et al. \cite{liu2018contextual} proposed a Contextual Outlier INterpretation (COIN) framework to explain anomalies spotted by detectors.
Goodall et al. \cite{goodall2019situ} presented "Situ," a streaming anomaly detection and visualization system for discovering and explaining suspicious behavior in computer network traffic and logs. The explanation is based on visualizing the context of the anomaly.
Collaris et al. \cite{collaris2018instance} designed two dashboards that provide explanations for fraud detected by a random forest algorithm. The explanations are based on existing explanation algorithms, such as the instance-level feature importance method \cite{palczewska2014interpreting}, partial dependence plots \cite{friedman2001greedy}, and local rule extraction (a variation of LIME \cite{ribeiroshould}). Arp et al. \cite{arp2014drebin} presented a method for the detection of Android malware using SVM and explain the decisions made by identifying the features that most contribute to a detection and checking whether the extracted features which contribute to the detection match common malware characteristics.
None of the above studies explained anomalies revealed by autoencoders or used SHAP values for the explanations as we have done in this work.

We are only aware of two recent studies that are related to ours, that are relevant both for explaining anomalies and unsupervised models. One of them compared Shapley values to reconstruction errors of features in principal component analysis (PCA) for explaining anomalies \cite{takeishi2019shapley}. In their research they changed a feature to make the instance anomalous and then compared between shapley values and reconstruction errors, while in our research we used shapley values of features to explain features with high reconstruction error.
The other study explains network anomalies from variational autoencoders \cite{nguyen2019gee}. Unlike our method, which treats the autoencoder as a black-box and provides explanations for the anomalies, they explained the anomalies by analyzing the gradients to identify the main features that affect the anomaly. We chose to work with a black-box explanation model because it has the advantage of being able to explain any autoencoder without knowledge of the exact model; in our case, just the input and output are known. 

\subsection{Explanations evaluation}
There is little consensus on the exact definition of interpretability in the context of  machine learning  and how to evaluate it. 
Doshi and Kim \cite{doshi2017roadmap} suggested three types of evaluations: (1) Application-grounded - Real humans, real tasks, (2) Human-grounded - Real humans, simplified tasks, (3) Functionally-grounded - No humans, proxy tasks.
Although it would be ideal to evaluate our explanations using domain experts who could assess the effectiveness of the explanation for them, such experts are not always available and do not always know specifically what they want to investigate (in terms of which instances may be anomalous and where to start their investigation of a potentially anomalous instance) before using a prototype \cite{liu2017towards}.
Some studies suggested categories to evaluate the explanations. Gunning \cite{gunning2017explainable} divided the explanation effectiveness into five categories: mental model, task performance, trust assessment, correctability, and user satisfaction.
Melis and Jaakkola \cite{melis2018towards} used three criteria for evaluation:
\begin{enumerate}
  \item Explicitness/Intelligibility: Are the explanations immediate and understandable?
  \item Faithfulness: Are relevance scores indicative of "true" importance?
  \item Stability: How consistent are the explanations for similar/neighboring examples?
\end{enumerate}

Lundberg and Lee \cite{lundberg2018consistent} evaluated their explanation algorithm by creating two simple decision trees so they knew which features contributed to a prediction.
Yang and Kim \cite{yang2019bim} recently presented a benchmark for prediction evaluation called BIM (Benchmark Interpretability Method). To construct this benchmark, they added noise objects to images and trained a classifier, without expecting that the noise objects would play a role in explaining the classification. 
There are a few studies (\cite{shrikumar2017learning, lundberg2017unified, lundberg2018consistent, samek2017evaluating}) that evaluate prediction explanations by changing the features that explained the most predicted class with the goal of reducing the probability for the predicted class or even changing the predicted class.

\section{Explaining autoencoder anomalies}
Our challenge was to design a method for explaining an anomaly revealed by an autoencoder (unsupervised), unlike existing explainability methods which are used to explain a prediction (supervised). We use an autoencoder to detect anomalies through the reconstruction error (anomaly score). Instances with high anomaly score are considered anomalous. We explain an anomaly score, which is the difference (error) between the input value and the output (reconstructed) value. An anomaly, if it exists, resides in the values of the input, and the explanatory model needs to explain why this instance was not predicted (reconstructed) well. Thus, an explanation must be connected to the error. Our method therefore computes the SHAP values of the reconstructed features and relates them to the true (anomalous) values in the input. 

Given input instance $X$ with a set of features $x_1, x_2,...,x_n$ and its corresponding output $X'$ and reconstructed values $x'_1, x'_2, ..., x'_n$, using an autoencoder model $f$, the reconstruction error of the instance is the sum of errors of each feature $L(X,X') = \sum_{i=1}^{n}(x_{i}-x'_{i})^2$. Let $x_{(1)},..., x_{(n)}$ be a reordering of the features in $errorList$, such that $|x_{(1)}- x'_{(1)}| \ge \dots \ge \ |x_{(n)}-x'_{(n)}| $, $topMfeatures = x_{(1)}, ..., x_{(m)}$ contains a set of features for which the total corresponding errors 
$topMerrors: |x_{(1)}- x'_{(1)}| ,..., |x_{(m)}- x'_{(m)}|$ represent an adjustable percent of $L(X,X')$.
Our method explains using SHAP values which features affected each of the high reconstruction errors in $topMfeatures$. 
\begin{algorithm}[h!]
\caption{Calculate SHAP values for $topMfeatures$}\label{SHAP values for top errors}
\begin{flushleft}
\textbf{Input:} 
$X$ - An instance we want to explain, $X1..j$ - Instances that kernel SHAP uses as background examples, $ErrorList$ - An ordered list of error per feature, $f$ - autoencoder model\linebreak 
\textbf{Output:} $shaptopMfeatures$ - SHAP values for each feature within $topMfeatures$
\end{flushleft}
\begin{algorithmic}[1]
\State $topMfeatures \gets$ top values from $ErrorList$
\ForAll{$i \in topMfeatures$}
\State $explainer \gets shap.KernelExplainer(f, X1..j)$
\State $shaptopMfeatures[i] \gets explainer.shapvalues($X$,$i$)$
\EndFor
\State return $shaptopMfeatures$
\end{algorithmic}
\end{algorithm}
Algorithm~\ref{SHAP values for top errors} presents the pseudocode for the process. First, we extract the features with the highest reconstruction error from the $ErrorList$ and save them in the $topMfeatures$ list. Next, for each feature $x'_i$ in $topMfeatures$, we use Kernel SHAP to obtain the SHAP values, i.e, the importance of each feature $x_1, x_2,..., x_n$ (except for $x_i$) in predicting the examined feature $x'_i$. Kernel SHAP receives $f$ and a background set with $j$ instances for building the local explanation model and calculating the SHAP values. Then, $f$ takes $X$ and $i$ as input and predicts $X'$; the value in the i'th feature (a feature in the $topMfeatures$) is returned by Algorithm~\ref{f(X,i)}. The result of this step is a two-dimensional list $shaptopMfeatures$, in which each row represents the SHAP values for one feature from the $topMfeatures$.
\begin{algorithm}[h!]
\caption{f(X,i) - predict the i'th feature}\label{f(X,i)}
\begin{flushleft}
\textbf{Input:} 
$X$ - An instance we want to explain, $i$ - The feature we want to get the prediction for (from the output vector of the autoencoder), $f$ - Autoencoder model\\
\textbf{Output:} $x'_i$ - The value of the i'th feature in $X'$ 
\end{flushleft}
\begin{algorithmic}[1]
\State $x'_i \gets f.predict($X$)[i]$
\State return $x'_i$
\end{algorithmic}
\end{algorithm}

We divide the SHAP values into values \textit{contributing} to the anomaly - those pushing the predicted (reconstructed) value away from the true value, and values \textit{offsetting} the anomaly - those pushing the predicted value towards the true value. The pseudocode in Algorithm~\ref{Explaining by contributing and offsetting SHAP values} presents how the SHAP values are divided. For each  feature (line 1), we check if the true (input) feature value is greater than the predicted value (line 2); the \textit{contributing} features are the features with a negative SHAP value (line 3), and the \textit{offsetting} features are the positives (line 4). If the predicted feature value is greater than the actual (input) value (line 5), then the \textit{contributing} features are the features with a positive SHAP value, and the \textit{offsetting} features are the negatives. This algorithm returns two lists, $shapContributing$ and $shapOffsetting$, that contain the \textit{contributing} and \textit{offsetting} features, along with their SHAP values, for each of the $topMfeatures$.
\begin{algorithm}[h!]
\caption{Divide SHAP values into those contributing and offsetting an anomaly}\label{Explaining by contributing and offsetting SHAP values}
\begin{flushleft}
\textbf{Input:} 
$shaptopMfeatures$ - SHAP values for each feature, $X$ - An instance we want to explain, $X'$ - The prediction for $X$\\
\textbf{Output:} $shapContribute$, $shapoffset$
\end{flushleft}
\begin{algorithmic}[1]
\ForAll{$i \in shaptopMfeatures$}
\If{$x_i > x'_i$}
\State $shapContribute[i] \gets shaptopMfeatures[i]<0$
\State $shapOffset[i] \gets shaptopMfeatures[i]>0$
\Else
\State $shapContribute[i] \gets shaptopMfeatures[i]>0$
\State $shapOffset[i] \gets shaptopMfeatures[i]<0$
\EndIf
\EndFor
return $shapContribute$, $shapOffset$
\end{algorithmic}
\end{algorithm}

\begin{figure}[h!]
    \centering
  \includegraphics[width=\textwidth,height=\textheight,keepaspectratio]{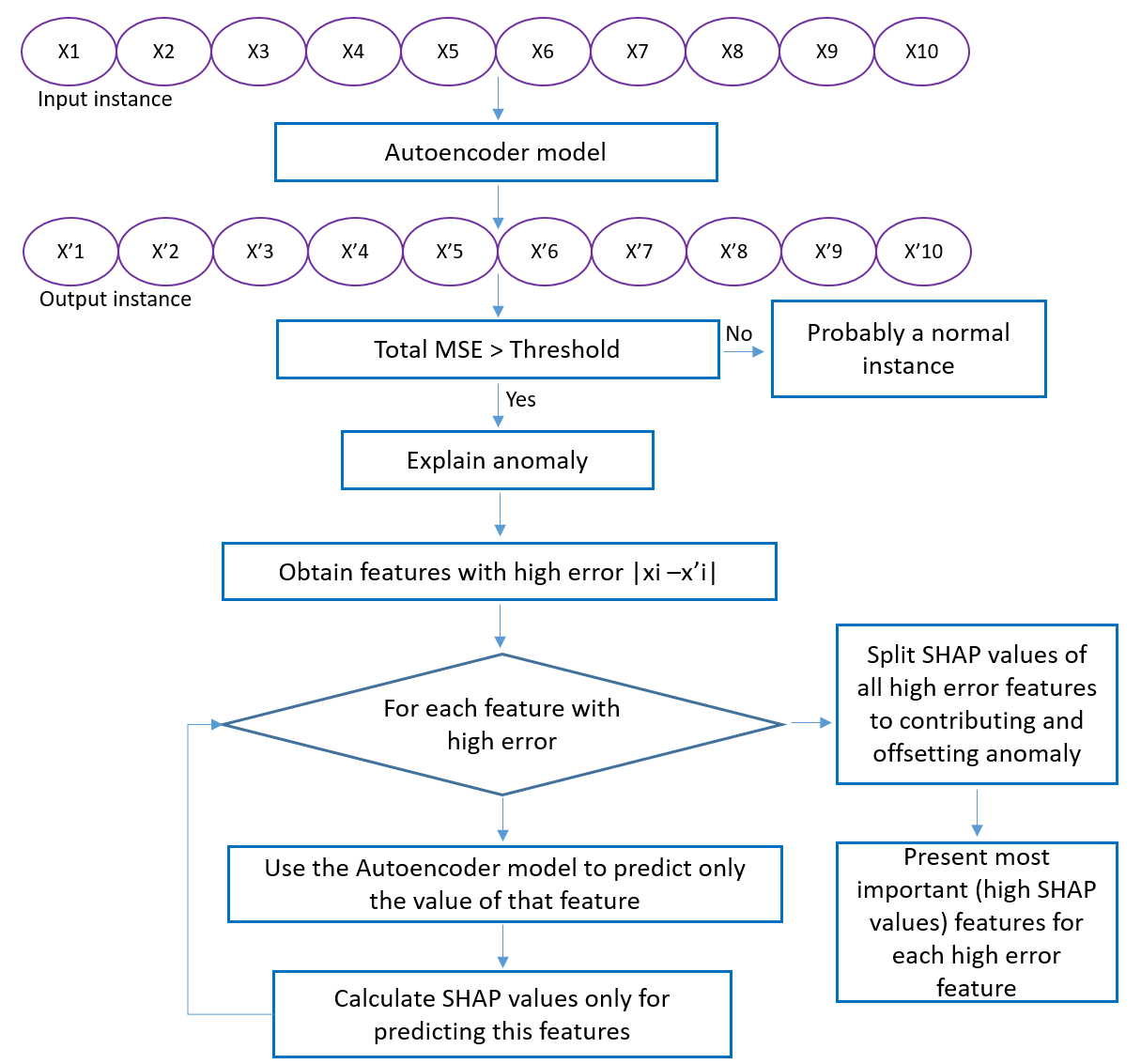}
  \caption{Flow chart describing the process of providing an explanation for an anomaly revealed by an autoencoder.}
  \label{fig:flow}
\end{figure}

The next step is selecting the features with high SHAP values of each of the features in the $topMfeatures$ list; so from each row in $shapContributing$ and $shapOffsetting$, we extract the highest values. Since our goal is to help the domain expert understand why an instance is an anomaly, we present the explanation in the form of a table that depicts the \textit{contributing} and \textit{offsetting} anomaly features, using colors which correspond to the SHAP values (Figure~\ref{fig:figure5}). A higher SHAP value (depicted by a darker color) means that the feature is more important for the prediction (\textit{contributing} features appear in red and \textit{offsetting} features appear in blue). The flow chart describing the process of providing an explanation for an anomaly revealed by an autoencoder can be seen in figure ~\ref{fig:flow}. The code for the explanation method can be found in github\footnote{https://github.com/ronniemi/explainAnomaliesUsingSHAP}.

\section{Example} \label{Example}
\begin{figure}[!tbp]
\centering
  \subfloat[Drug amount ($X_{3}$), days between prescription and purchase date ($X_{6}$), and doctor name ($X_{8}$) are the features with the highest errors]{\includegraphics[width=0.45\textwidth]{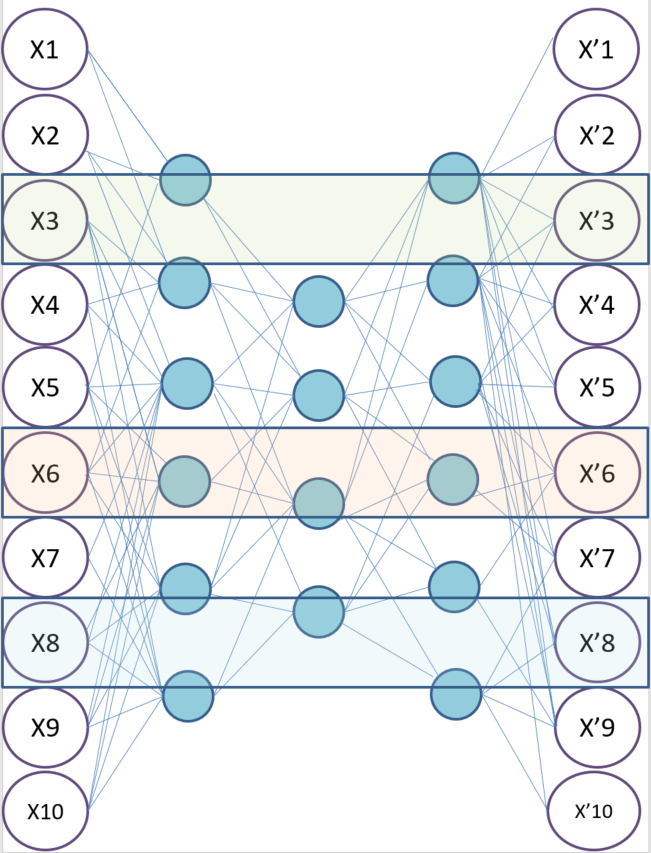}\label{fig:figure2}}
  \hfill
  \subfloat[Calculating contributing and offsetting features for drug amount]{\includegraphics[width=0.45\textwidth]{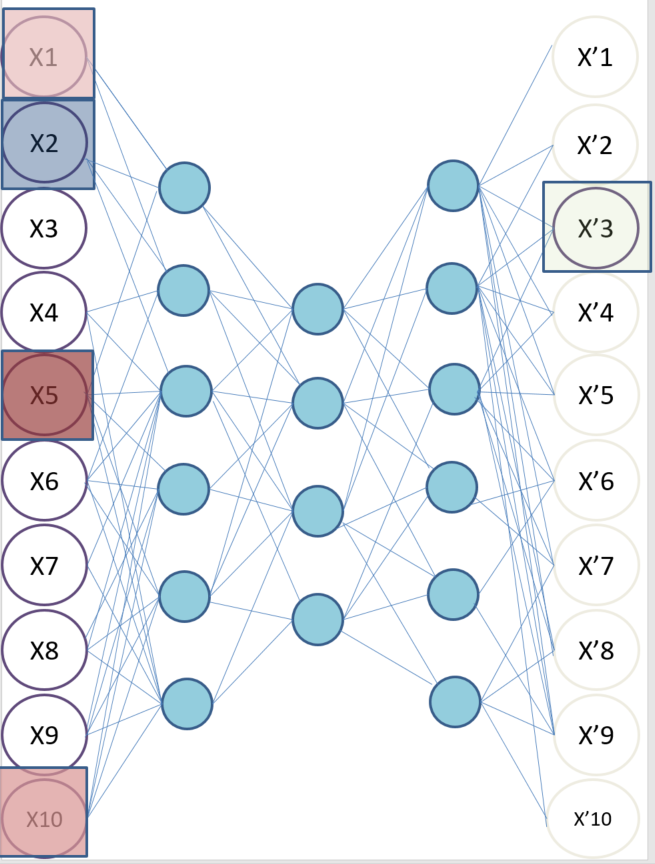}\label{fig:figure3}}
 \caption{Explaining the drug amount ($X_{3}$) error.}
\end{figure}

To demonstrate our method, we use an example in which we are trying to detect drug abuse using a prescription database. Each record has ten features that may point to drug abuse. The instance presented in Figure~\ref{fig:figure2}, which has a high reconstruction error, is a prescription for a large amount of painkillers prescribed to a 30-years-old man who has no comorbidities but was recently involved in a car accident.

\textbf{Extracting top error features.}
Since the total reconstruction error is calculated from the error of each feature ($|X_{i}-X'_{i}|$), we can extract the features with the highest reconstruction error. Let's assume that features $X_{3}$ (drug amount), $X_{6}$ (days between prescription and purchase date), and $X_{8}$ (doctor name) have the highest reconstruction errors; therefore, these are the features that we explain using SHAP.

\textbf{Calculating SHAP values for a feature with high error.}
In order to explain the reconstruction error in the drug amount feature ($X_{3}$), we use the autoencoder to predict the value of the drug amount $X'_{3}$, as in Figure~\ref{fig:figure3}, and use SHAP to obtain the importance of each feature in the network in predicting $X'_{3}$, relative to a baseline which is calculated using the background set, as in Algorithm~\ref{SHAP values for top errors}.

\textbf{Features contributing or offsetting an anomaly.}
Figure~\ref{fig:figure4} presents a plot with positive (depicted in blue in the figure) and negative (red) SHAP values. Assume that the real value of feature $X_{3}$ is one, and the autoencoder predicted that $X'_{3}$ equals 0.01. To divide the features based on whether they \textit{contribute} or \textit{offset} the anomaly, we use the true (input) value, output (reconstructed) value, and the polarity of the SHAP values, as in Algorithm~\ref{Explaining by contributing and offsetting SHAP values}. Only event=car accident (${X_2}$) pushed the value towards the true value, \textit{offsetting} the anomaly, while time from last prescription=five days (${X_5}$), age=30 (${X_{10}}$), and medical background=no disease (${X_1}$) pushed the value away from the true value towards the prediction, \textit{contributing} to the anomaly. Because the young patient had no comorbidities and requested painkillers five days before, the autoencoder predicted that the amount should be much lower than what was prescribed. Perhaps the event feature ($X_2$) offsets the anomaly, because the fact that the patient was involved in a car accident makes this prescription correct.

\textbf{Depiction of contributing and offsetting anomaly features.}
Figure~\ref{fig:figure5} shows how we visually depict the features \textit{contributing} and \textit{offsetting} the anomaly to the domain expert. For each feature in the $topMfeatures$ ($X'_{3}$,$X'_{6}$,$X'_{8}$), we show the \textit{contributing} features in red. For example, $X_{5}$ is the feature that contributed most to the error of feature $X_{3}$. In the third column we show the real value of that feature. Then we present the features that offset the anomaly in blue; the last column contains the real value of that feature.

\section{The motivation for using our method for explaining anomalies.}
The prescription described in the example in section \ref{Example} may be normal. Painkillers are commonly prescribed, even for young, healthy people. So why is it anomalous? 
Without using our suggested method to provide an explanation for the anomalies revealed by the autoencoder, the domain expert would receive an alert regarding this prescription, and the only clue he/she would have about its anomalous nature is the list of features with the highest reconstruction errors. In this example the clues would be the features of drug amount, days between prescription and purchase date, and doctor name. Without the explanations of the features that are most important in affecting the reconstruction error, the reason for the anomaly remains vague. Using our method, we are able to explain what affected the incorrect predictions (and thus the reconstruction errors). 

When the expert sees the features that most affected the prediction (the features that explain the reconstruction error), it is clearer why the autoencoder detected it as an anomaly. In this example we had only ten features, but in many real-world problems, the number of features is much higher, which makes it much harder to understand the anomaly without a proper explanation.

\begin{figure}[h!]
    \centering
  \includegraphics[width=0.9\textwidth]{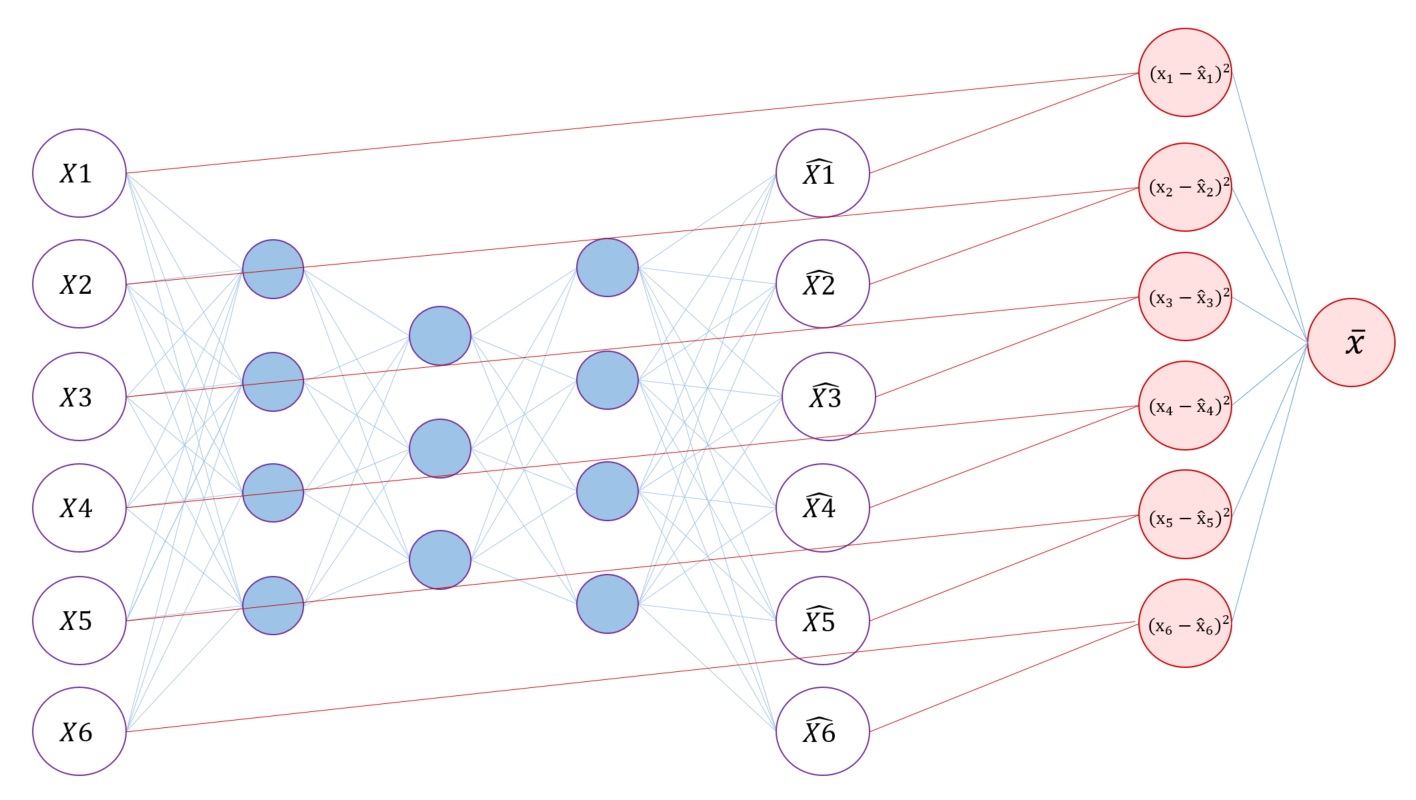}
  \caption{Adding a layer of the total reconstruction error to provide explanations for the total reconstruction error using SHAP.}
  \label{fig:michael}
\end{figure}

\begin{figure}[!tbp]
\centering
  \subfloat[Explaining $X'_{3}$] {\includegraphics[width=0.45\textwidth]{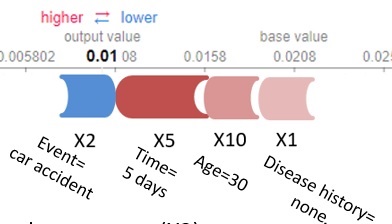}\label{fig:figure4}}
    \hfill
  \subfloat[Features that contribute and offset the anomaly]{\includegraphics[width=0.45\textwidth]{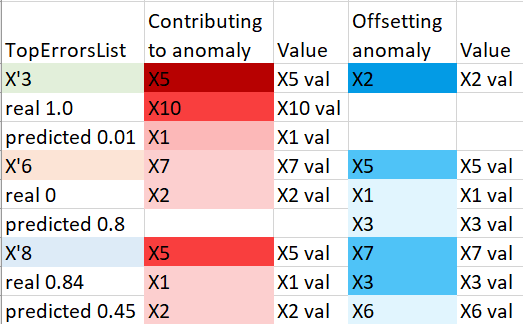}\label{fig:figure5}}
 \caption{(a) Highest SHAP values in explaining drug amount ($X'_{3}$). The real value of $X_{3}$ is one, so features $X_{5}$,$X_{10}$, and $X_{1}$ push the value of $X'_{3}$ away from the true value, \textit{contributing} to the anomaly. Feature $X_{2}$ pushes the value towards the true value of $X_{3}$, \textit{offsetting} the anomaly. (b) A table containing explanations for all features with high error.
 }
\end{figure}

Another way of using SHAP to explain anomalies is to add another layer to the autoencoder after it is trained, in such a way that each neuron represents the reconstruction error of one feature, and then add an output layer that sums the reconstruction errors of all of the features, as in figure ~\ref{fig:michael}.

In this case we change the explanation task to a regression task and use Kernel SHAP without adaptations, in order to obtain the features that are the most important in affecting the total reconstruction error of the instance, rather than obtaining the features that are most important in a prediction for each feature with high reconstruction error separately. 
We conducted tests using this approach and found that in most of the cases, the features with the highest SHAP values were the same as the features with the highest reconstruction errors. These findings strengthened our confidence in our decision to use SHAP for explanations and based on this positive finding, we continued with our method because it provides a more comprehensive explanations to the experts by focusing on the connection between the features with high reconstruction error and the features that are most important in affecting the reconstruction error.

\section{Evaluation}
We evaluated our suggested method for explaining anomalies using four different approaches: (1) we performed a user study conducted on real data with domain experts, (2) we used simulated data in which we know which features should explain the anomalies, (3) we assessed the robustness of the explanations on real-world data, and (4)  We examined the affect of changing the value of the features that explain the anomaly on the anomaly score.

\subsection{Datasets}\label{datasets}
We preformed the evaluations on the four datasets described below. Different evaluations were performed on different datasets. 
\subsubsection{Warranty claims} \label{warranty}
This unsupervised dataset contains 15,000 warranty claims from a large car manufacturer with 1,000 features per instance. To detect anomalies we trained an autoencoder and provided a list of 114 claims with high anomaly scores to domain experts for inspection. The threshold for choosing the top anomalies was determined using the interquartile range (IQR) measure. The top anomalies are the ones for which $Reconstruction error > Q3 + (Q3 - Q1)*1.5)$. After the domain experts examined the top anomalies, we received feedback regarding which of the claims were in fact interesting anomalies.
\subsubsection{Artificial dataset} \label{artificial datasets}
We generated an artificial dataset that consists of one million instances with six features, using the following procedure: The first four features, $X_{1}$, $X_{2}$, $X_{3}$ and $X_{4}$, were generated randomly with feature values between zero and one. 

The following two features are a linear combination of some of the four previous features: $X_{5}$=$X_{1}$+$X_{2}$, $X_{6}$=$X_{3}$+$X_{4}$. Each such linear combination yields two other linear combinations of the features: $X_{1} = X_{5} - X_{2}$ and $X_{2} = X_{5} - X_{1}$. The same applies to features $X_{3}$, $X_{4}$, and $X_{6}$.
Then, we randomly selected 5,000 records to create anomalies by changing $X_{5}$ or $X_{6}$ to a random value between zero and one as in Table~\ref{tab:anomalies}.

\begin{table}[]
\caption{Nine examples of the artificial dataset and anomalies (generated anomalies in bold): class zero - normal cases, class one - generated anomalies (the value of $X_{5}$ or $X_{6}$ does not fit the linear combination).}  
\label{tab:anomalies}
\scriptsize
\begin{adjustbox}{center}
\begin{tabular}{|c|c|c|c|c|c|c|}
\hline
$X_{1}$  & $X_{2}$  & $X_{3}$  & $X_{4}$  & $X_{5}$          & $X_{6}$          & Class \\ \hline
0.447 & 0.608 & 0.445 & 0.869 & \textbf{0.34} & 1.314         & 1     \\ \hline
0.499 & 0.481 & 0.386 & 0.862 & 0.98          & 1.248         & 0     \\ \hline
0.05  & 0.287 & 0.275 & 0.264 & 0.336         & 0.538         & 0     \\ \hline
0.808 & 0.031 & 0.811 & 0.546 & 0.838         & 1.356         & 0     \\ \hline
0.323 & 0.798 & 0.025 & 0.961 & 1.121         & \textbf{0.24} & 1     \\ \hline
0.074 & 0.962 & 0.184 & 0.159 & 1.035         & 0.344         & 0     \\ \hline
0.104 & 0.061 & 0.189 & 0.383 & 0.164         & 0.572         & 0     \\ \hline
0.931 & 0.542 & 0.747 & 0.905 & \textbf{0.66} & 1.652         & 1     \\ \hline
0.336 & 0.275 & 0.051 & 0.493 & 0.611         & 0.544         & 0     \\ \hline
\end{tabular}
\end{adjustbox}
\end{table}

\begin{figure}[h!]
\centering
  \subfloat[Model 1: $X_{1} + X_{2}=X_{5}$, $X_{3} + X_{4}=X_{6}$]{\includegraphics[width=0.45\textwidth]{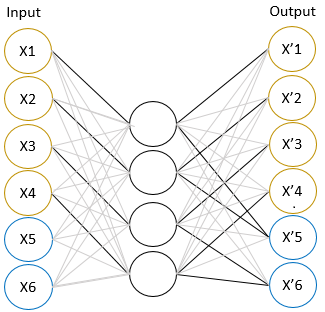}\label{fig:perfect3}}
    \hfill
  \subfloat[Model 2: $X_{2} =X_{5}-X_{1}$, $X_{4} =X_{6}-X_{3}$]{\includegraphics[width=0.45\textwidth]{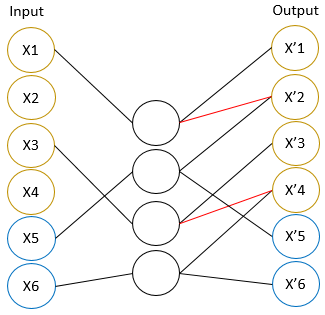}\label{fig:perfect4}}
    \hfill
  \subfloat[Model 3: $X_{1} =X_{5}-X_{2}$, $X_{3} =X_{6}-X_{4}$]{\includegraphics[width=0.45\textwidth]{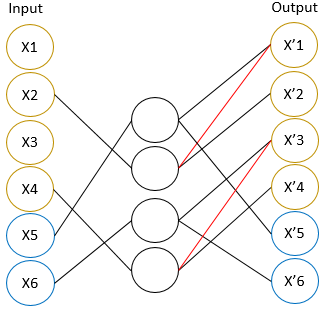}\label{fig:perfect5}}
 \caption{Perfect autoencoders with weights matching the linear combinations (black lines=weight of one, red lines=weight of minus one, grey lines=weight of zero).}
\end{figure}
\subsubsection{KDD Cup 1999}\label{kdd}
The KDD Cup 1999\footnote{http://kdd.ics.uci.edu/databases/kddcup99/kddcup99.html} dataset, obtained from the UCI Machine Learning Archive, is a well-known anomaly detection dataset of intrusions simulated in a military network environment. It has been widely used in many studies. We use the ’10 percent’ version of the data which is a subset of the data that contains records classified as 22 different types of attacks. For the evaluation of our method, we ignored 21 of these types of attacks and used only the most frequent attack after the normal class. Specifically, we used the “back” attack which is a type of denial-of-service attack, resulting in a binary dataset containing 97,278 normal connections (class zero) and 2,203 ‘back’ attack connections (class one). This dataset contains 41 features. 
\subsubsection{Credit Card Fraud}\label{credit}
The Credit Card Fraud Detection dataset was obtained from Kaggle.\footnote{https://www.kaggle.com/mlg-ulb/creditcardfraud} This dataset contains credit card transactions from two days in September 2013 for European cardholders; in this dataset, there are 492 fraudulent transactions out of a total of 284,807 transactions. This data contains 30 numerical features which are the result of a  preprocess PCA transformation. Due to confidentiality issues, we have no information about the meaning of the features, but the given features, V1, V2, V3 ... V28, are principal components obtained with PCA. ‘Time’ and ‘Amount’ are the only features that have not been transformed with PCA. Feature 'Class' is the label - one in case of fraud and zero otherwise.

\subsection{User study}
As part of a project aimed at developing an anomaly detection method for monitoring the cost of the warranty claims of a large car manufacturer, we developed an autoencoder-based anomaly detector. The detection of fraud or human error is part of an effective cost monitoring process which is extremely important to the company, enabling them to reduce costs, improve their products, and better serve their customers.
Until now, the domain experts at the company have produced reports based on predefined rules (according to KPIs) to reveal irregularities in warranty claims.
The output of the autoencoder-based anomaly detector revealed anomalies that the domain experts were unable to detect using the existing process. However, an explanation of the anomalies was needed in order to convince the domain experts of the correctness of the anomalies found. 
In order to accomplish this we used an autoencoder to detect anomalies from 15,000 warranty claims, with 1,000 features. Domain experts received a list of the 114 top anomalies with the visual depiction for explaining the anomalies that we provided.
They were instructed to decide whether the anomaly should be inspected further, using their existing system and the visual depiction.
We conducted interviews with four of the experts after the experiment, and they reported that the explanations provided a clear indication of how to examine the anomalies (more specifically, they noted that by using the visual depiction they were able to focus first on the most important explanatory \textit{contributing} features (depicted in a darker color) to examine the anomaly). 

We note that in over 85 percent of the claims, the first one to three explanatory features explained why the claim was anomalous. The domain experts also said that our method helped them handle complicated claims more than it helped them with common, easy claims. It took a day for them adapt to working with the visual depiction tool, but after that, they were comfortable with these visualizations and satisfied with the explanations.

\subsection{Correctness of explanations}
To evaluate the correctness of our method of explaining anomalies detected by an autoencoder using SHAP, we built "perfect" autoencoders for which we know the relations between the features and thus have a ground truth to explain the anomaly. The effectiveness of our method is demonstrated when the anomalies are explained using the known correct set of features.
For this evaluation we did not split the features into those that contribute and offset the anomaly, since we were interested in examining all of the explanatory features.
This approach is similar to the evaluation approach used in \cite{lundberg2018consistent}, where they created two simple decision trees so they would know which features are attributed to a prediction. We used the same idea and built autoencoders with known connections, so that when there is an anomaly, we would know which features explain it. The dataset used for this evaluation is artificial, as explained in ~\ref{artificial datasets}.

\subsubsection{Creating perfect autoencoders}\label{perfect} 
We manually built autoencoders for each linear combination mentioned above in ~\ref{artificial datasets}, so they can produce an output identical to the input. 
The autoencoders include an input layer, an inner layer and output layers. For each autoencoder, we set the weights from the input layer to the inner layer to one for all of the independent features and to zero for the dependent features. The inner neuron that gets the value of the independent feature $X_{i}$ will have a weight of one or minus one to $X'_{i}$ in the output layer, depending on the linear combination of the features.
If the linear combination is $X_{5}$=$X_{1}$+$X_{2}$, then $X_{1}$ and $X_{2}$ are the independent features. Therefore, the weights from $X_{1}$ and $X_{2}$ in the input layer to the matching neurons in the inner layer will be one. The weights from those inner neurons to $X'_{1}$ and $X'_{2}$ will also be one. $X'_{5}$ in the output layer will get a weight of one from the neurons that received the value of $X_{1}$ and $X_{2}$ in the inner layer, to match the linear combination $X_{5}$=$X_{1}$+$X_{2}$. 

On the other hand, if the relationship is $X_{1}$=$X_{5}$-$X_{2}$, then $X_{5}$ and $X_{2}$ are the independent features. Therefore, the weights from $X_{5}$ and $X_{2}$ in the input layer to the matching neuron in the inner layer will be one. The weights from those inner neurons to $X'_{5}$ and $X'_{2}$ will be one. $X'_{1}$ in the output layer will receive a weight of one from the neuron that received the value of $X_{5}$ in the inner layer, and a weight of minus one from the neuron that received the value of $X_{2}$ in the inner layer, to match the linear combination $X_{1}$=$X_{5}$-$X_{2}$. Some examples for the perfect autoencoders are presented in Figures ~\ref{fig:perfect3}, ~\ref{fig:perfect4}, and ~\ref{fig:perfect5}. 

\textbf{Results.} 
Since we know the dependencies of the different features of this artificial dataset, we expect that our explanation model will provide the correct explanation for an anomaly. For example, if $X_{1}$ is 0.35 and $X_{2}$ is 0.24, then $X_{5}$ should be equal to 0.59, but if this record is identified as an anomaly, $X_{5}$ will randomly receive any value between zero and one. In the autoencoder that represents the linear combination $X_{1}+X_{2}=X_{5}$, we expect that $X_{5}$ will have a high reconstruction error, and as a result, this record will be considered an anomaly. We also expect that the explanation model based on SHAP will provide an explanation for the anomaly that includes only features $X_{1}$ and $X_{2}$ (they will obtain high SHAP values), meaning that these two features explain the reconstruction error of $X_{5}$. In the autoencoder that represents the linear combination $X_{1} =X_{5}-X_{2}$, we expect that $X_{1}$ will have a high reconstruction error, and the explaining features for the anomaly will be $X_{5}$ and $X_{2}$. Ultimately, we are interested in the set of explanatory features, which are the combination of the features with a high reconstruction error and the features that explain the reconstruction error (the features with the highest SHAP values that contribute or offset the anomaly). The set of explanations we expect to see for an anomaly caused by changing the value of $X_{5}$ should be equal to all three linear combinations, as seen in Table~\ref{tab:explanations}. We ran our explanation method on 5,000 anomalies, and for each anomaly, we examined whether the set of explanatory features matched our expectations, for all three models. Only seven anomalies were not explained well (one feature was missing), and after investigating these results, we realized that the background set chosen for calculating the SHAP values was responsible for the mismatch. When we increased the size of the background set, all anomalies were explained exactly as expected. This points out the need to examine how to optimally choose the background set, which we plan to do in future research.
\begin{center}
    \begin{table}
      \caption{Features with high reconstruction errors, features that explain the high reconstruction error (have high SHAP values), and the set of explanatory features. This table represents an example of the set of features explaining an anomaly caused by the change of $X_5$ (without the second linear combination).}
      \label{tab:explanations}
      \scriptsize
      \begin{adjustbox}{center}
      \centering
      \begin{tabular}{|c|c|c|c|}
      \hline
       Model   & \begin{tabular}[c]{@{}c@{}}High reconstruction\\ error features\end{tabular} & \begin{tabular}[c]{@{}c@{}}Explanatory\\ features\end{tabular} & \begin{tabular}[c]{@{}c@{}}Set of features explaining\\ the anomaly\end{tabular} \\ \hline
       Model 1 & $X_{5}$ & $X_{1}$, $X_{2}$ & {$X_{1}$, $X_{2}$, $X_{5}$} \\ \hline
        Model 2 & $X_{2}$ & $X_{5}$, $X_{1}$ & {$X_{1}$, $X_{2}$, $X_{5}$} \\ \hline
        Model 3 & $X_{1}$ & $X_{5}$, $X_{2}$ & {$X_{1}$, $X_{2}$, $X_{5}$} \\ \hline
     \end{tabular}
    \end{adjustbox}
    \end{table}
\end{center}

\subsubsection{Creating a more complex perfect autoencoders}
We also tried to use our perfect autoencoder method on binary data; to do this, we generated 1,000,000 records with four independent features of random binary values. Next, we created two dependent features that are the output of logical “AND” and “OR” operations of the independent binary features. The first dependent feature $X_{5}$ is the result of an “AND” operation between $X_{1}$ and $X_{2}$. The second dependent feature $X_{6}$ is the result of an “OR” operation between $X_{3}$ and $X_{4}$. Eventually we obtained four independent binary features and two dependent binary features. In order to add complexity to the model, we added 16 independent binary features, so we have a total of 22 binary features. Then, we added noise to 15,000 random records by changing the output of one of the dependent binary features $X_{5}$ or $X_{6}$. 

In this experiment, instead of manually setting the weight to zeros and ones as we did in the previous experiment on the perfect autoencoder, we allowed the autoencoder to train its weights on the binary data generated. 

\textbf{Results.} In all of the anomalies we created, the set of explanatory features matched our expectations, meaning it was the correct set of explanatory features.

\subsection{Robustness of the explanation model}
We developed a procedure to evaluate the robustness of the explanation model, inspired by the Benchmark Interpretability Method (BIM) presented recently by Yang and Kim \cite{yang2019bim}. To construct this benchmark, they added noise objects to images and trained a classifier, without expecting that the noise objects would play a role in explaining the classification. 
To evaluate our model's robustness, we created noise features within the dataset and observed the use of these features in the explanations of anomalies when our suggested method is used, as explained earlier, with SHAP and  LIME \cite{ribeiro2016should}. In this aspect of our evaluation, we aim to show that the explanations obtained when using SHAP are more robust to noise than those obtained when using LIME.
\subsubsection{Creating the explanation feature set}
Similar to the way we examined the "perfect" autoencoder (described in ~\ref{perfect}), we need to define the $Explanatory Feature Set$ for the robustness evaluation.
To create the $Explanatory Feature Set$, we first need to define two parameters:
$Reconstruction Error Percent$ and $SHAP values Selection$.

\textbf{$Reconstruction Error Percent$.}
This parameter is the percent of the total reconstruction error we want to explain. When we visually depict the output of our method, we chose to show 80 percent of the reconstruction error. This value was obtained through trial and error on the Warranty Claims dataset. This value affects the number of features we explain, and therefore it affects the number of features we receive in the final explanatory set for a specific record. In addition, this parameter affects the run time of our method; as its value increases, we will explain a larger portion of the reconstruction error, i.e., calculate SHAP values to explain the reconstruction error of more features.

\textbf{$SHAP values Selection$.}
This parameter represents the method for selecting the features used to explain the feature with the reconstruction error (i.e., the anomaly that needs to be explained).
The selected features are added to $Explanatory Feature Set$. We used the following feature selection methods:
\begin{enumerate}
	\item Selecting features with a SHAP value higher than the mean SHAP value
	\item Selecting features with a SHAP value higher than the median SHAP value
	\item Selecting the five features with the highest SHAP values 
\end{enumerate}
The combination of $Reconstruction Error Percent$ and $SHAP values Selection$ creates the $Explanatory Feature Set$ in the following way:
The features whose reconstruction errors sum to $Reconstruction Error Percent$ of the total reconstruction error of the instance are selected. Each such feature is added to the $Explanatory Feature Set$, along with the set of features that explain it that meet the $SHAP values Selection$ criteria. The order of the $Explanatory Feature Set$ is set so that each explained feature that appears is followed by the selected features that explain it.
In this process of creating the explanatory feature list we may encounter features that already appear on the list. For example, one explained feature might explain another feature's error. In that case, a feature will appear in the  $Explanatory Feature Set$ only once.

\textbf{Example.} We received the errors in Table \ref{tab:reconstructionerrors} for a given record with a total error of 2.97. Let's assume that $Reconstruction Error Percent$=0.5; in this case, we explain only the first two features $F_1$, $F_2$ because 50 percent of the total error is 1.485, and we can see that the cumulative error of the top two features represents more than 50 percent of the total reconstruction error. Next, we use SHAP to explain the reconstruction errors of $F_1$ and $F_2$. The explanations for the reconstruction errors of $F_1$ and $F_2$ are shown in \ref{tab:f1}. For this example, $SHAP values Selection$ are features higher than the mean SHAP value.  The mean SHAP value of the four explanatory features we obtain in the explanation of $F_1$ is 0.4. We can see that only the first two features have a SHAP value greater than 0.4, so we add only $F_2$ and $F_5$ to $Explanatory Feature Set$, after the explained feature $F_1$. The mean SHAP value of the four explanatory features we get in the explanation of $F_2$ is 0.285, so only the first feature  $F_3$ has a SHAP value greater than 0.285 and thus is added to $Explanatory Feature Set$ (after the explained feature $F_2$). 

The final explanatory feature set after removing duplicates is ${F_1, F_2, F_5, F_3}$ (the order is important for calculating the robustness measure).
The same process of creating $Explanatory Feature Set$ is performed when examining the explanations of our suggested method using LIME instead of SHAP.
\begin{center}
    \begin{table}
      \caption{Reconstruction error of features}
      \label{tab:reconstructionerrors}
      \scriptsize
      \begin{adjustbox}{center}
      \begin{tabular}{|c|c|c|}
      \hline
       Feature name & Reconstruction error & Cumulative error\\ \hline
        $F_1$ & $0.95$ & $0.95$\\
        \hline
        $F_2$ & $0.83$ & $1.78$\\
        \hline
        $F_3$ & $0.52$ & $2.3$\\
        \hline
        $F_4$ & $0.43$ & $2.73$\\
        \hline
        $F_5$ & $0.24$ & $2.97$\\
       \hline
     \end{tabular}
    \end{adjustbox}
    \end{table}
\end{center}

\begin{center}
    \begin{table}
      \caption{Features that explain the reconstruction errors of $F_1$ and $F_2$.}
      \label{tab:f1}
      \scriptsize
      \begin{adjustbox}{center}
      \begin{tabular}{|c|c|c|}
        \hline
        Explained feature & Explanatory features & SHAP value\\ 
        \hline
        $F_1$ & $F_2$ & $0.7$\\
        \hline
        $ $ & $F_5$ & $0.5$\\
        \hline
        $ $ & $F_3$ & $0.3$\\
        \hline
        $ $ & $F_4$ & $0.1$\\
        \toprule
        $F_2 $ & $F_3$ & $0.7$\\
        \hline
        $ $ & $F_4$ & $0.25$\\
        \hline
        $ $ & $F_1$ & $0.16$\\
        \hline
        $ $ & $F_5$ & $0.03$\\
       \hline
     \end{tabular}
     \end{adjustbox}
    \end{table}
\end{center}
\subsubsection{Robustness test process}
We randomly chose features to become noise features. The values of each such feature were changed randomly to a value within the original range of the feature but from a uniform distribution.
Then, we built an autoencoder using the normal (not anomalous) instances.
After training the autoencoder, we explained the anomalies from the dataset using our explanation method with kernel SHAP. We then used again our proposed method but with LIME instead of SHAP and compared between the two $Explanatory Feature Set$. We expect to see as few explanations that include the noise feature as possible. If the noise feature exists in the $Explanatory Feature Set$, we expect it to be located at a low position, meaning that it is not very important in the explanation. 
Since we change the noise features to a random value, we performed this process five times for each of the explanation methods (SHAP and LIME) for each of the noise features. It is important to note that in each iteration $k=1..5$, we examine the explanations provided by SHAP and LIME on the same data. 
\subsubsection{Sensitivity analysis}
To evaluate our method, we preformed sensitivity analysis. We explained each anomaly using our method with $Reconstruction Error Percent$=[0.1, 0.2, 0.3, 0.4, 0.5, 0.6, 0.7, 0.8, 0.9] and for each value of $Reconstruction Error Percent$, we used all of the possible $SHAP values Selection$ methods. Each combination of the values of $Reconstruction Error Percent$ and $SHAP values Selection$ created different $Explanatory Feature Set$.
\subsubsection{Evaluation criterion}
\begin{figure}[!tbp]
\centering
  \subfloat[Explaining the five features with the highest SHAP values]{\includegraphics[width=0.45\textwidth]{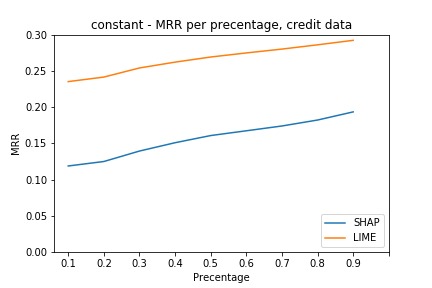}\label{fig:creditconstant}}
  \hfill
  \subfloat[Explaining features higher than the mean]{\includegraphics[width=0.45\textwidth]{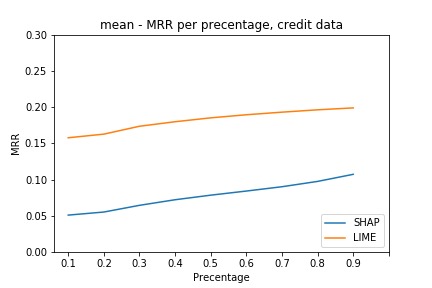}\label{fig:creditmean}}
  \subfloat[Explaining features higher than the median]{\includegraphics[width=0.45\textwidth]{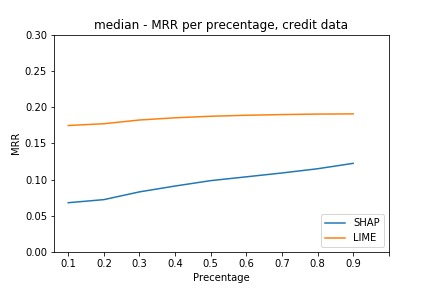}\label{fig:creditmedian}}
 \caption{Credit Card dataset - MRR of SHAP vs. LIME. The x-axis presents the explanation set size, and the y-axis presents the MRR. Each graph represents a different way of creating the set, as explained earlier.}\label{fig:mrrcredit}
\end{figure}

\begin{figure}[!tbp]
\centering
  \subfloat[Explaining the five features with the highest SHAP values]{\includegraphics[width=0.45\textwidth]{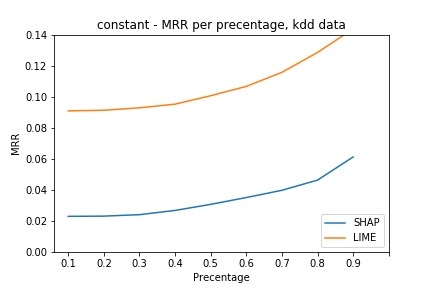}\label{fig:kddconstant}}
  \hfill
  \subfloat[Explaining the features higher than the mean]{\includegraphics[width=0.45\textwidth]{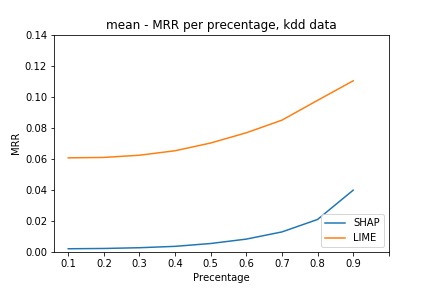}\label{fig:kddmean}}
  \subfloat[Explaining the features higher than the median]{\includegraphics[width=0.45\textwidth]{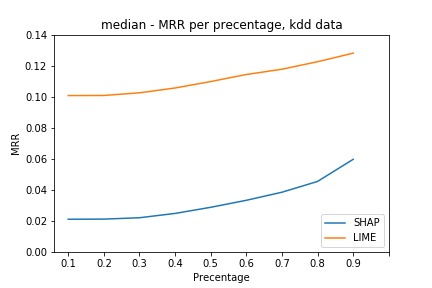}\label{fig:kddmedian}}
 \caption{KDD Cup '99 dataset - MRR of SHAP vs. LIME. The x-axis presents the explanation set size, and the y-axis presents the MRR. Each graph represents a different way of creating the set, as explained earlier.}\label{fig:mrrkdd}
\end{figure}

We measure the performance of this evaluation method using the mean reciprocal rank (MRR) measure which considers the position of the noise feature in the explanatory feature set; the higher the noise feature appears in the explanatory feature set, the higher the value of the MRR. This means that we want the MRR value to be as low as possible.
The order of operations in the process are as follows:
\begin{enumerate}
	\item Choose a random feature (the noise feature)
	\item Change its values to random values that distribute uniformly
	\item Train the autoencoder model on the data that contains the noise feature with new random values
	\item Explain anomalous instances with the suggested method, once using kernel SHAP and once with LIME, and calculate mean MRR for each one of them.
\end{enumerate}
\textbf{Results.} 
For each noise feature and $Explanatory Feature Set$ (with its two parameters, $Reconstruction Error Percent$ and $SHAP values Selection$), we calculated the mean MRR of SHAP and LIME, and then we used a paired t-test to determine their significance. Figures ~\ref{fig:mrrcredit} and ~\ref{fig:mrrkdd} show the graphs of the MRR for the different explanation sets, based on the $Reconstruction Error Percent$ and $SHAP values Selection$ values, for the two datasets.
The mean MRR of SHAP is significantly lower ($p-value < 0.001$) than the mean MRR of LIME for all of the explanation sets on both datasets. This means that LIME tends to use the noise features for explanations more, so it is less robust than SHAP.

\subsection{Effectiveness of explanations in reducing anomaly score}
In the literature, there are a few works (\cite{shrikumar2017learning, lundberg2017unified, lundberg2018consistent, samek2017evaluating}) that evaluate prediction explanations by changing the features that most explain the predicted class, with the goal of changing the predicted class. 
We evaluate our method using the same idea but adapt it to anomalies and an unsupervised model. The autoencoder provides an anomaly score for an instance, which is the sum of the reconstruction errors of all of the features. Our method explains the anomaly using the features with the highest reconstruction error and the features that explain those features. In this evaluation, we change the values of the first feature in $Explanatory Feature Set$ and the features that explain the first feature.
We also chose to evaluate the method in this way given the way the experts  used the explanations in the user study. They examined the anomaly using the explanations, starting with the feature with the highest reconstruction error and its explanatory features, and then moved on to the next feature. This means that they expect that the features at the beginning of the $Explanatory Feature Set$ will help them the most in understanding the cause of the anomaly.
The $Explanatory Feature Set$ is built in three ways: 
\begin{enumerate}
    \item Explain anomalous instances with our method using kernel SHAP 
    \item Explain anomalous instances with our method using LIME
    \item Creating a set of random features
\end{enumerate}
We changed the values of the features that exist in the $Explanatory Feature Set$, expecting that this change would decrease the total MSE (reconstruction error). To examine the impact of changing the anomaly feature values on the reconstruction error, we need to examine different ways of changing these values:
\begin{enumerate}
	\item Replacing values with mean values – we calculate the mean values of each of the features in the dataset and change the feature value to its mean value.
	\item Replacing a value with its predicted (reconstructed) value by the trained autoencoder. 
\end{enumerate}
\textbf{Results.} 
Table ~\ref{tab:reducemse} shows (for two datasets) the average MSE (anomaly score) of 400 anomalies before changes, the average MSE after changing the value of the feature with the highest reconstruction error (to a mean value and a predicted value), and the MSE after changing the values of the explanatory features of the feature that needs to be explained, with all of the explanation sets (SHAP, LIME, and a random set); for each method, the value was changed to a mean value and a predicted value.
We checked for statistical significance using a one-way ANOVA with repeated measures, with $p-value < 0.05$, followed by a paired t-test, with confidence level $p-value < 0.05$, between the methods ($Explanatory Feature Set$ built using SHAP, LIME and a random set of features) and the different value change options (mean, predicted). All tests were performed with vectors of 400 anomaly scores. The tests were done using a few configurations: (1) with the same parameter configuration for all the methods (value changes to a mean value, and another test for all methods with value changes to the predicted value), (2) with the best parameter configuration for each method (lowest average reconstruction error). 

Our results indicate that for both datasets, we were able to significantly reduce the anomaly score when using the explanation set created with SHAP to change the feature values to the mean value or the predicted value, more than when using the explanation set created with LIME or a random set of features.
\begin{table}[]
\caption{Comparison of MSE reduction by changing values of features that explain that anomalies.}
     \label{tab:reducemse}
\scriptsize
\begin{tabular}{|c|c|c|c|c|c|c|c|c|c|}
\hline
\multirow{2}{*}{Data} & 
\multirow{2}{*}{Mean MSE} & 
\multicolumn{2}{c|}{\begin{tabular}[c]{@{}c@{}}Mean MSE after change\\ feature with highest error \end{tabular}} &  
\multicolumn{2}{c|}{SHAP} & \multicolumn{2}{c|}{LIME} & \multicolumn{2}{c|}{Random} \\ \cline{3-10} &       & mean & predicted                                  & mean & predicted    
& mean & predicted    
& mean & predicted  \\ \hline
Credit                & 0.0252                    & 0.0197                                                & 0.0197                                                  &\bfseries 0.0173     &\bfseries 0.0173       & 0.0189     & 0.0187       & 0.0181      & 0.0182        \\ \hline
KDD                   & 0.0005                    & 0.0011                                                & 0.0005                                                  & \bfseries0.0006     &\bfseries 0.0003       & 0.0023     & 0.0004       & 0.0017      & 0.0004        \\ \hline
\end{tabular}
\end{table}

\section{Discussion}
In this paper, we presented an approach for explaining anomalies identified by an autoencoder using SHAP. We examined the correctness of the method on an artificial dataset and perfect autoencoders. In the process, we realized that in order to evaluate our method in other ways, we needed to create an explanation set. While this set is not as informative to domain experts as the visual depiction we provide, where they can better understand which features were most important in creating a high reconstruction error, it is necessary for evaluating the set of explanatory features; furthermore, the order of the features within the set is important to our evaluation process.

With those things in place, we evaluated the robustness of the explanations and the effectiveness of the explanations in reducing the anomaly score on real-world data using the $Explanatory Feature Set$ that was built using SHAP as suggested in this paper but also using LIME, for comparison. We assessed the robustness of the method by changing each time one feature to "noise" feature and thus should not explain an anomaly. We checked if and where the explanation methods use the noise feature in the explanation set. We found that SHAP uses the noise feature significantly less and at a lower position in the explanation set than LIME. This evaluation was done using different methods of creating the explanation sets.
The effectiveness of the explanations in reducing the anomaly score was similarly assessed on both SHAP and LIME, but we also used a random feature set as a baseline. In this evaluation, we changed the first features in the explanation set to other values (mean value, predicted value, random value) and examined the change in the anomaly score for all three methods. The idea behind this evaluation was that if the features in the explanation set explain the anomaly, then when we change them, the instance would be less anomalous, meaning its anomaly score would be reduced. In this evaluation we saw that SHAP reduces the reconstruction error more than LIME and the random features set.
It is worth mentioning the following insights regarding the suggested method:
\begin{enumerate}
	\item \textit{The goal of the explanation method:} We presented a method for explaining anomalies using (1) the features with the highest reconstruction errors from an autoencoder, and (2) the features that are most important in affecting the reconstruction error of the feature needs to be explained. The main aim of the research was to provide a more comprehensive explanation to the user by focusing on the connection between the features with high reconstruction error and the features that are most important in affecting the reconstruction error.  
	\item \textit{Background set:} Since this method treats the autoencoder model as a black-box, we need a background set to create a local explanation model. The background set in this research consists of 200 instances from the dataset, without any constraints or considerations. For a fair comparison, the same background set was used with both SHAP and LIME. Choosing an appropriate background set can lead to different explanation models. We are currently addressing this important issue in another study.
	\item \textit{Visualization:} In this research, we presented a visual depiction of the explanation, as part of our joint work with a large car manufacturer. Further research should be performed to explore other ways of presenting the explanations. We believe that the form the presentation takes will likely depend on the needs of the company and user, as well as other other factors.
\end{enumerate}

\section{Conclusion and future work}
We developed a method that uses SHAP values which are based on game theory to explain anomalies revealed by an autoencoder. The feedback obtained from the domain experts about the explanations for the anomalies was positive. In our evaluation of the explanation method using perfect autoencoders, we showed that the set of the features that have high reconstruction errors, together with the features that contribute or offset the anomaly, are what explains an anomaly. The evaluation of the proposed method's robustness showed that creating an $Explanatory Feature Set$ as part of our explanation method using SHAP is more robust than creating the set using LIME.
We plan to further develop the proposed method by examining the background set used for the explanation model and evaluate the explanation method with more complicated autoencoders and additional datasets. 
\label{}

\bibliography{mybibfile}

\begin{thebibliography}{10}
\expandafter\ifx\csname url\endcsname\relax
  \def\url#1{\texttt{#1}}\fi
\expandafter\ifx\csname urlprefix\endcsname\relax\def\urlprefix{URL }\fi
\expandafter\ifx\csname href\endcsname\relax
  \def\href#1#2{#2} \def\path#1{#1}\fi

\bibitem{miller2018explanation}
T.~Miller, Explanation in artificial intelligence: Insights from the social
  sciences, Artificial Intelligence (2018).

\bibitem{shortliffe1975model}
E.~H. Shortliffe, B.~G. Buchanan, A model of inexact reasoning in medicine,
  Mathematical biosciences 23~(3-4) (1975) 351--379.

\bibitem{ribeiroshould}
M.~T. Ribeiro, S.~Singh, C.~Guestrin, “why should i trust you?” explaining
  the predictions of any classifier (2016).

\bibitem{kindermans2017reliability}
P.-J. Kindermans, S.~Hooker, J.~Adebayo, M.~Alber, K.~T. Sch{\"u}tt,
  S.~D{\"a}hne, D.~Erhan, B.~Kim, The (un) reliability of saliency methods,
  stat 1050 (2017) 2.

\bibitem{lundberg2017unified}
S.~M. Lundberg, S.-I. Lee, A unified approach to interpreting model
  predictions, in: Advances in Neural Information Processing Systems, 2017, pp.
  4765--4774.

\bibitem{ribeiro2016should}
M.~T. Ribeiro, S.~Singh, C.~Guestrin, Why should i trust you?: Explaining the
  predictions of any classifier, in: Proceedings of the 22nd ACM SIGKDD
  international conference on knowledge discovery and data mining, ACM, 2016,
  pp. 1135--1144.

\bibitem{shrikumar2017learning}
A.~Shrikumar, P.~Greenside, A.~Kundaje, Learning important features through
  propagating activation differences, in: Proceedings of the 34th International
  Conference on Machine Learning-Volume 70, JMLR. org, 2017, pp. 3145--3153.

\bibitem{erfani2016high}
S.~M. Erfani, S.~Rajasegarar, S.~Karunasekera, C.~Leckie, High-dimensional and
  large-scale anomaly detection using a linear one-class svm with deep
  learning, Pattern Recognition 58 (2016) 121--134.

\bibitem{paula2016deep}
E.~L. Paula, M.~Ladeira, R.~N. Carvalho, T.~Marzag{\~a}o, Deep learning anomaly
  detection as support fraud investigation in brazilian exports and anti-money
  laundering, in: Machine Learning and Applications (ICMLA), 2016 15th IEEE
  International Conference on, IEEE, 2016, pp. 954--960.

\bibitem{sakurada2014anomaly}
M.~Sakurada, T.~Yairi, Anomaly detection using autoencoders with nonlinear
  dimensionality reduction, in: Proceedings of the MLSDA 2014 2nd Workshop on
  Machine Learning for Sensory Data Analysis, ACM, 2014, p.~4.

\bibitem{rumelhart1985learning}
D.~E. Rumelhart, G.~E. Hinton, R.~J. Williams, Learning internal
  representations by error propagation, Tech. rep., California Univ San Diego
  La Jolla Inst for Cognitive Science (1985).

\bibitem{bengio2007scaling}
Y.~Bengio, Y.~LeCun, et~al., Scaling learning algorithms towards ai,
  Large-scale kernel machines 34~(5) (2007) 1--41.

\bibitem{hinton2006reducing}
G.~E. Hinton, R.~R. Salakhutdinov, Reducing the dimensionality of data with
  neural networks, science 313~(5786) (2006) 504--507.

\bibitem{hinton2006fast}
G.~E. Hinton, S.~Osindero, Y.-W. Teh, A fast learning algorithm for deep belief
  nets, Neural computation 18~(7) (2006) 1527--1554.

\bibitem{goodfellow2016deep}
I.~Goodfellow, Y.~Bengio, A.~Courville, Deep learning, MIT press, 2016.

\bibitem{bertsimas2018interpretable}
D.~Bertsimas, A.~Orfanoudaki, H.~Wiberg, Interpretable clustering via optimal
  trees, arXiv preprint arXiv:1812.00539 (2018).

\bibitem{radev2004centroid}
D.~R. Radev, H.~Jing, M.~Sty{\'s}, D.~Tam, Centroid-based summarization of
  multiple documents, Information Processing \& Management 40~(6) (2004)
  919--938.

\bibitem{jolliffe2011principal}
I.~Jolliffe, Principal component analysis, Springer, 2011.

\bibitem{maaten2008visualizing}
L.~v.~d. Maaten, G.~Hinton, Visualizing data using t-sne, Journal of machine
  learning research 9~(Nov) (2008) 2579--2605.

\bibitem{bertsimas2017optimal}
D.~Bertsimas, J.~Dunn, Optimal classification trees, Machine Learning 106~(7)
  (2017) 1039--1082.

\bibitem{liu2000clustering}
B.~Liu, Y.~Xia, P.~S. Yu, Clustering through decision tree construction, in:
  Proceedings of the ninth international conference on Information and
  knowledge management, ACM, 2000, pp. 20--29.

\bibitem{lundberg2018consistent}
S.~M. Lundberg, G.~G. Erion, S.-I. Lee, Consistent individualized feature
  attribution for tree ensembles, arXiv preprint arXiv:1802.03888 (2018).

\bibitem{kauffmann2018towards}
J.~Kauffmann, K.-R. M{\"u}ller, G.~Montavon, Towards explaining anomalies: A
  deep taylor decomposition of one-class models, arXiv preprint
  arXiv:1805.06230 (2018).

\bibitem{amarasinghe2018toward}
K.~Amarasinghe, K.~Kenney, M.~Manic, Toward explainable deep neural network
  based anomaly detection, in: 2018 11th International Conference on Human
  System Interaction (HSI), IEEE, 2018, pp. 311--317.

\bibitem{liu2018contextual}
N.~Liu, D.~Shin, X.~Hu, Contextual outlier interpretation, in: Proceedings of
  the 27th International Joint Conference on Artificial Intelligence, AAAI
  Press, 2018, pp. 2461--2467.

\bibitem{goodall2019situ}
J.~R. Goodall, E.~D. Ragan, C.~A. Steed, J.~W. Reed, G.~D. Richardson, K.~M.
  Huffer, R.~A. Bridges, J.~A. Laska, Situ: Identifying and explaining
  suspicious behavior in networks, IEEE transactions on visualization and
  computer graphics 25~(1) (2019) 204--214.

\bibitem{collaris2018instance}
D.~Collaris, L.~M. Vink, J.~J. van Wijk, Instance-level explanations for fraud
  detection: A case study, arXiv preprint arXiv:1806.07129 (2018).

\bibitem{palczewska2014interpreting}
A.~Palczewska, J.~Palczewski, R.~M. Robinson, D.~Neagu, Interpreting random
  forest classification models using a feature contribution method, in:
  Integration of reusable systems, Springer, 2014, pp. 193--218.

\bibitem{friedman2001greedy}
J.~H. Friedman, Greedy function approximation: a gradient boosting machine,
  Annals of statistics (2001) 1189--1232.

\bibitem{arp2014drebin}
D.~Arp, M.~Spreitzenbarth, M.~Hubner, H.~Gascon, K.~Rieck, C.~Siemens, Drebin:
  Effective and explainable detection of android malware in your pocket., in:
  Ndss, Vol.~14, 2014, pp. 23--26.

\bibitem{takeishi2019shapley}
N.~Takeishi, Shapley values of reconstruction errors of pca for explaining
  anomaly detection, in: 2019 International Conference on Data Mining Workshops
  (ICDMW), IEEE, 2019, pp. 793--798.

\bibitem{nguyen2019gee}
Q.~P. Nguyen, K.~W. Lim, D.~M. Divakaran, K.~H. Low, M.~C. Chan, Gee: A
  gradient-based explainable variational autoencoder for network anomaly
  detection, in: 2019 IEEE Conference on Communications and Network Security
  (CNS), IEEE, 2019, pp. 91--99.

\bibitem{doshi2017roadmap}
F.~Doshi-Velez, B.~Kim, A roadmap for a rigorous science of interpretability,
  stat 1050 (2017) 28.

\bibitem{liu2017towards}
S.~Liu, X.~Wang, M.~Liu, J.~Zhu, Towards better analysis of machine learning
  models: A visual analytics perspective, Visual Informatics 1~(1) (2017)
  48--56.

\bibitem{gunning2017explainable}
D.~Gunning, Explainable artificial intelligence (xai), Defense Advanced
  Research Projects Agency (DARPA), nd Web (2017).

\bibitem{melis2018towards}
D.~A. Melis, T.~Jaakkola, Towards robust interpretability with self-explaining
  neural networks, in: Advances in Neural Information Processing Systems, 2018,
  pp. 7786--7795.

\bibitem{yang2019bim}
M.~Yang, B.~Kim, Bim: Towards quantitative evaluation of interpretability
  methods with ground truth, arXiv preprint arXiv:1907.09701 (2019).

\bibitem{samek2017evaluating}
W.~Samek, A.~Binder, G.~Montavon, S.~Lapuschkin, K.-R. M{\"u}ller, Evaluating
  the visualization of what a deep neural network has learned, IEEE
  transactions on neural networks and learning systems 28~(11) (2017)
  2660--2673.

\end{thebibliography}

\end{document}